\documentclass[runningheads]{llncs}
\usepackage{lmodern}
\usepackage{graphicx}
\usepackage{array}
\usepackage{booktabs}
\usepackage{amsmath}
\usepackage{amssymb}
\usepackage{marvosym}

\begin{document}

\title{Deepfake Forensics Adapter: A Dual-Stream Network for Generalizable Deepfake Detection}
\titlerunning{Deepfake Forensics Adapter}
%
%

\author{Jianfeng Liao\inst{1} \and
Yichen Wei\inst{1} \textsuperscript{(\scalebox{1.8}{\Letter})} \and
Raymond Chan Ching Bon\inst{2} \and
Shulan Wang\inst{1} \and
Kam-Pui Chow\inst{3} 
\and Kwok-Yan Lam\inst{4}
}

\authorrunning{J. Liao et al.}
%
\institute{Shenzhen Technology University, Guangdong, China 
\email{202300204010@stumail.sztu.edu.cn}, \email{weiyichen@sztu.edu.cn}, \email{wangshulan@sztu.edu.cn}
\\
\and
Singapore Institute of Technology, Singapore, Singapore 
\email{raymond.chan@singaporetech.edu.sg}
\\
\and
The University of Hong Kong, Hong Kong, China\\
\email{chow@cs.hku.hk}
\\
\and
Nanyang Technological University, Singapore, Singapore \\
\email{kwokyan.lam@ntu.edu.sg}
}
\maketitle              
%
\let\thefootnote\relax
\footnotetext{This work was supported by the Natural Science Foundation of Top Talent of SZTU(grant no. GDRC202414) and the Guangdong Provincial Engineering and Technology Research Center for Universities (grant no. 2024GCZX004).}

\begin{abstract}
The rapid advancement of deepfake generation techniques poses significant threats to public safety and causes societal harm through the creation of highly realistic synthetic facial media. While existing detection methods demonstrate limitations in generalizing to emerging forgery patterns, this paper presents Deepfake Forensics Adapter (DFA), a novel dual-stream framework that synergizes vision-language foundation models with targeted forensics analysis. Our approach integrates a pre-trained CLIP model with three core components to achieve specialized deepfake detection by leveraging the powerful general capabilities of CLIP without changing CLIP parameters: 1) A Global Feature Adapter is used to identify global inconsistencies in image content that may indicate forgery, 2) A Local Anomaly Stream enhances the model's ability to perceive local facial forgery cues by explicitly leveraging facial structure priors, and 3) An Interactive Fusion Classifier promotes deep interaction and fusion between global and local features using a transformer encoder. Extensive evaluations of frame-level and video-level benchmarks demonstrate the superior generalization capabilities of DFA, particularly achieving state-of-the-art performance in the challenging DFDC dataset with frame-level AUC/EER of 0.816/0.256 and video-level AUC/EER of 0.836/0.251, representing a 4.8\% video AUC improvement over previous methods. Our framework not only demonstrates state-of-the-art performance, but also points out a feasible and effective direction for developing a robust deepfake detection system with enhanced generalization capabilities against the evolving deepfake threats. Our code is available at \url{https://github.com/Liao330/DFA.git}

\end{abstract}
\section{Introduction} 
The rapid advancement of deepfake techniques, such as Generative Adversarial Networks (GANs)~\cite{label_1,label_2,label_3} and Diffusion Models~\cite{label_4,label_5}, has enabled the generation of synthetic images and videos with unprecedented realism, often indistinguishable from authentic content to human observers. Historical precedents underscore these dangers: since late 2017, when an anonymous Reddit user ("deepfakes") pioneered the illicit insertion of celebrity faces into pornographic content, misuse cases have increased in sophistication and impact~\cite{label_36}. Recent incidents, such as the 2024 Hong Kong corporate fraud involving deepfake video impersonation (resulting in HK\$200 million losses)~\cite{label_35} and South Korea’s "Nth Room 2.0" scandal, exemplify the misuse of the technology for large-scale harm~\cite{label_37}. Such synthetic media, often featuring manipulated facial attributes, now permeate illegal activities ranging from disinformation campaigns and digital evidence tampering to non-consensual pornography. Law enforcement agencies, including the FBI~\cite{label_6} and Europol~\cite{label_45}, have issued urgent warnings, with Europol characterizing deepfakes as “one of the most direct, tangible and destructive applications of Artificial Intelligence.” Despite these concerns, conventional forensics methods, whether analyzing signal-level artifacts (e.g., double JPEG compression), physical inconsistencies, or semantic anomalies (e.g., metadata discrepancies)~\cite{label_7,label_8,label_9,label_10,label_11}, demonstrate diminishing reliability against evolving deepfake techniques.

Contemporary deepfake detection predominantly employs deep learning-based binary classifiers operating at image or video levels. Image-centric approaches leverage convolutional neural networks (CNNs), including ResNeXt~\cite{label_38}, ConvNet~\cite{label_39}, and Xception~\cite{label_12}, to assess frame-level authenticity by capturing subtle spatial artifacts, with Xception excelling in trace detection through separable depthwise convolutions. Video-level methods extend this by integrating temporal analysis: spatial features extracted via CNNs are processed sequentially using architectures such as Vision Transformers (ViT) to identify cross-frame inconsistencies~\cite{label_13,label_14,label_15}.  However, despite progress, these methods struggle with generalization when encountered in synthetic media from novel generators (e.g., advanced GAN variants or emerging diffusion models). This limitation highlights the necessity for detection systems with enhanced cross-domain adaptability. Recent breakthroughs in Foundation Models~\cite{label_16,label_17} particularly Contrastive Language-Image Pre-training (CLIP) model~\cite{label_17}, offer promising solutions through their zero-shot inference capabilities and task adaptation potential. While CLIP-based linear probing achieves state-of-the-art performance in general synthetic image detection~\cite{label_18}, its application to facial deepfake detection (a domain requiring precise localization of facial anomalies) remains underexplored, which presents both methodological challenges and novel opportunities.

Building on recent breakthroughs in adapting large foundation models, we propose the Deepfake Forensics Adapter (DFA), a dual-stream framework that leverages CLIP’s pre-trained knowledge through innovative adapter modules. Realizing that facial forgery often introduces anomalies in specific semantic regions, such as the eyes and mouth, we employ a skillfully designed Global feature Adapter that interacts with CLIP. This interaction generates attention biases to guide the focus of CLIP'S attention toward discriminative features. Simultaneously, a Local Anomaly Stream uses facial structural information to extract local features oriented towards key regions. Finally, an Interactive Fusion Classifier effectively integrates features derived from both the global context and the local feature, enabling robust detection of complex forgery patterns.

\begin{figure}[h] 
    \centering 

    \includegraphics[width=0.8\linewidth]{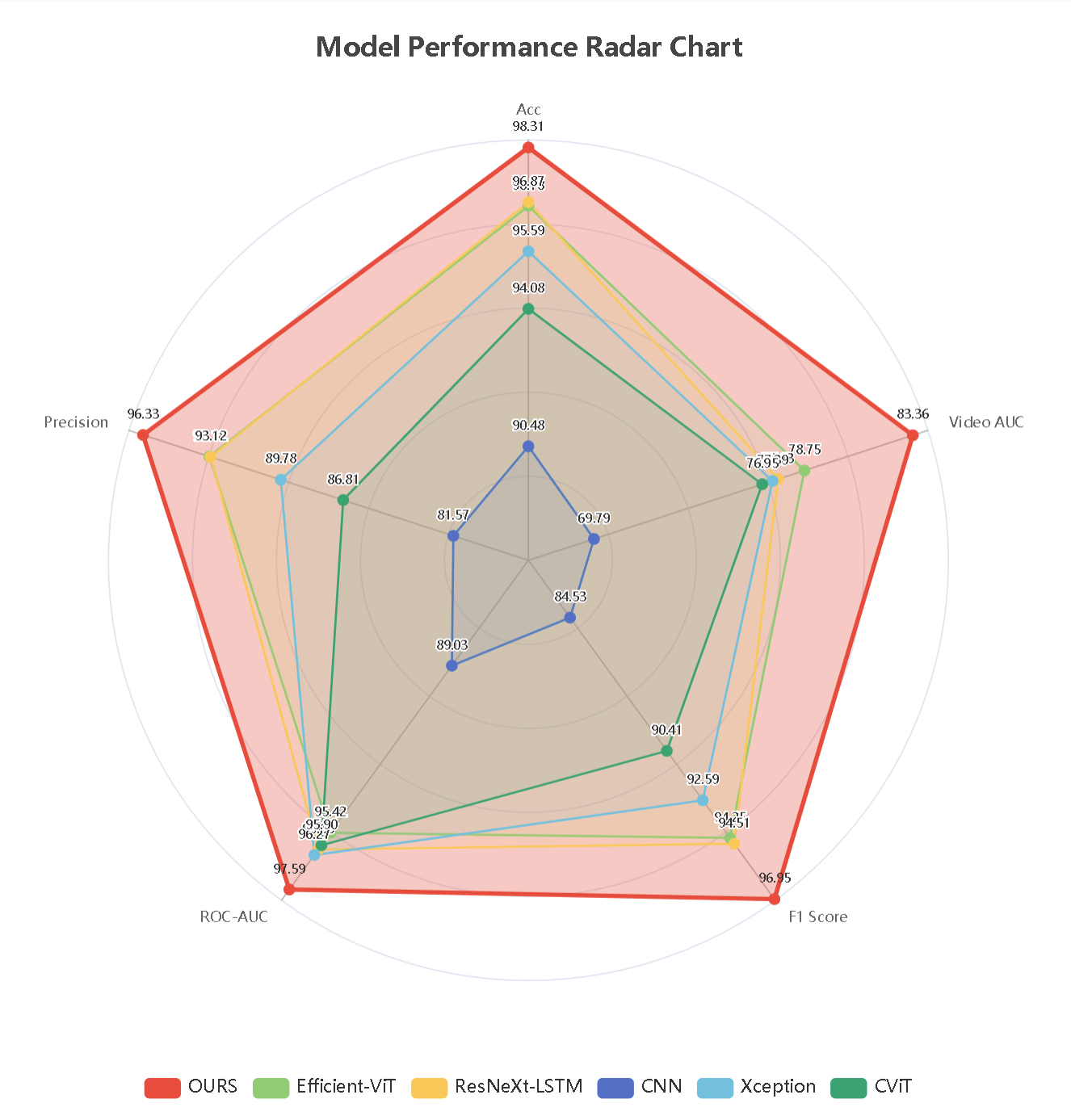} 

    \caption{The radar chart shows how different models perform across several metrics.} 

    \label{fig:figure_1} 
\end{figure}

The DFA framework operates through three interconnected components:
(1) A \textbf{Global Feature Adapter (Global)} integrates the visual features of the pre-trained model into the shallow layers of the adaptation process. Through an attention mechanism, it feeds knowledge acquired during adaptation back to the pre-trained model, guiding its focus onto forgery traces without altering the original parameters. (2) A \textbf{Local Anomaly Stream (Local)} leverages anatomical priors derived from facial landmarks to isolate and amplify inconsistencies in critical regions (e.g., irregular pupil geometries, asymmetric lip textures), addressing the limited localized perception of conventional methods. (3) An \textbf{Interactive Fusion Classifier (IFC) Module} receives feature tensors from both the Global and Local stream. It performs feature fusion to form a comprehensive forgery representation. Leveraging the interaction enhancement capabilities of Transformers, it captures dependencies between features, thereby enhancing sensitivity to forgery cues. Finally, the fused information is transformed into a binary classification output, providing an accurate real versus forged judgment. 
Through extensive experiments and ablation studies, the proposed method demonstrates superior performance compared to existing state-of-the-art methods under unified dataset evaluation protocols. The results, illustrated in Figure~\ref{fig:figure_1}, validate the effectiveness of our approach. 

Our main contributions are summarized as follows:
\begin{enumerate}
    \item \textbf{We propose a novel CLIP-based dual-stream adapter framework (DFA) that enhances generalization in deepfake detection.} By integrating the pre-trained CLIP model with customized adapter modules, our architecture preserves CLIP's original parameters while adapting its capabilities to facial forgery tasks through a dual-stream interaction strategy. This approach effectively leverages CLIP's semantic-visual knowledge while improving sensitivity to diverse manipulation artifacts. As a result, it achieves superior cross-dataset performance compared to existing methods.
    \item \textbf{We develop a novel local anomaly stream and interactive fusion mechanism to address localized forgery detection.} The Local stream utilizes facial structural priors to focus on critical regions (e.g., eyes, mouth), while the IFC employs transformer-based feature interaction to model dependencies between local anomalies and global context. This design overcomes the limitations of traditional methods in capturing subtle regional inconsistencies, significantly improving detection robustness.
    \item \textbf{We conduct extensive evaluations on multiple benchmarks to validate the effectiveness of DFA, including the challenging DFDC dataset. }Our framework achieves the state-of-the-art performance in the DFDC dataset, not only reaching 0.816 and 0.836 AUC at the frame level and video level, respectively, and the video-level performance is 4.8\% better than the existing methods, but also achieving the lowest EER at present, which verifies its high accuracy and good generalization ability in actual deployment.
\end{enumerate}

\section{Related Work}
\subsection{Deepfake Generation}
\label{sec:2_1}
The fast development of deepfake technology has spurred the emergence of various advanced generation methods. Although these techniques can generate highly realistic synthetic facial content, they also erode trust in cyberspace, threatening growth of the digital economy, and compromising individual privacy. This section explores the main current categories of deepfake generation techniques:

\subsubsection{Face Reenactment}
Face reenactment techniques aim to transfer dynamic features, such as facial expressions, lip movements, or head poses, from a source person to a target person's video. This enables the target person to mimic the source's actions and expressions while retaining their own identity. Early reenactment methods often relied on hand-designed feature point tracking or parametric models based on the Facial Action Coding System (FACS). In recent years, methods based on Generative Adversarial Network (GAN), such as DeepFaceLab~\cite{label_19} and the First Order Motion Model (FOMM)~\cite{label_20}, have significantly improved the realism and fluency of reenactment results by learning latent representations of facial motion.
\subsubsection{Face Swapping}
Face-swapping techniques involve replacing the entire facial region of a source person with the head area in a target person's image or video frame, thereby achieving an identity-level appearance transformation. The primary technical challenges of such methods include seamlessly blending the boundaries of the swapped face and maintaining consistent lighting conditions. Classic tools like Faceswap~\cite{label_21} typically use autoencoder architectures for the extraction and reconstruction of facial features. Newer methods, such as FSGAN~\cite{label_22}, introduce techniques like image segmentation guidance and style transfer to further improve the perceptual realism of the resulting swapped images. In response to such identity-level manipulations, specific detection strategies have been developed, such as identity-driven methods that focus on inconsistencies between the source and target identities~\cite{label_44}.
\subsubsection{Face Editing}
Face editing techniques focus on making local, attribute-level modifications to facial images, such as changing hairstyle, skin tone, perceived age, or adding/removing virtual accessories such as eyeglasses. GAN-based models, such as StarGAN~\cite{label_23} and StyleGAN~\cite{label_3}, enable the editing of multiple facial attributes through conditional inputs or manipulation within the latent space. In particular, StyleGAN and its subsequent variants~\cite{label_24}, leveraging their disentangled feature representation capabilities, can generate highly controllable editing results. However, ensuring the consistency of the latent information from the edited area with that of other areas remains a key challenge faced by these methods.
\subsubsection{Entire Face Synthesis}
The goal of all facial synthesis techniques is to generate completely fictional but visually highly realistic facial images, typically achieved by learning the feature distribution of large-scale facial datasets. StyleGAN2~\cite{label_24} and recent Diffusion Models, such as Stable Diffusion~\cite{label_5}, have achieved breakthrough progress in this domain. These deep learning models can capture the complex statistical patterns of facial features, thereby generating various virtual identities that do not exist in the real world. Although synthesized faces may be visually indistinguishable from real ones, they can sometimes lack certain underlying semantic consistencies (e.g., identity stability across multiple images), providing potential identification cues for semantic-based detection methods.

\begin{figure}[h] 

    \centering 

    \includegraphics[width=1\linewidth]{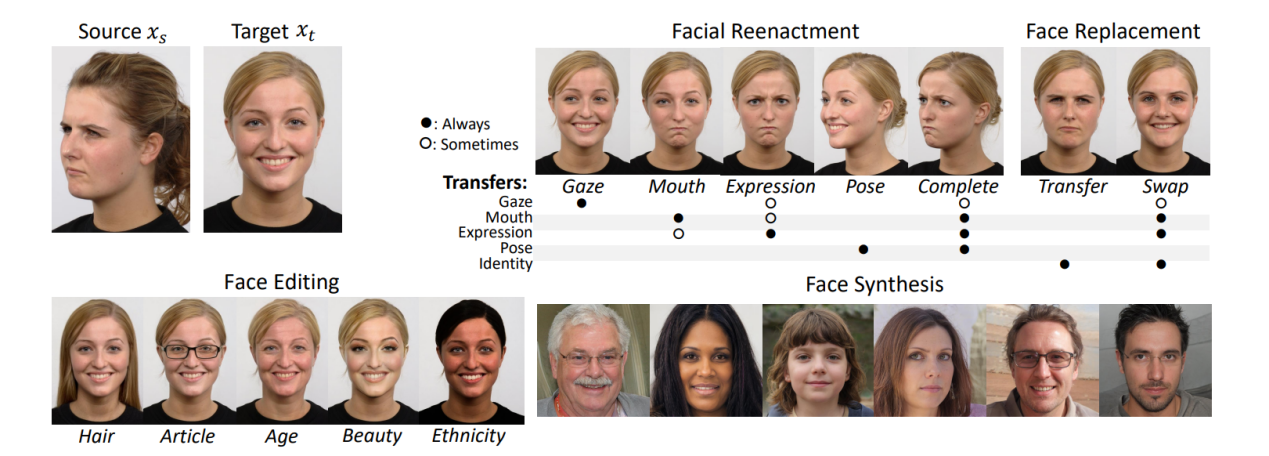} 

    \caption{Examples of reenactment, replacement, editing, and synthesis of deepfakes of the human face.~\cite{label_25}} 

    \label{fig:figure_2} 
\end{figure}

\subsection{Deepfake Detection}
\label{sec:2_2}
\subsubsection{Traditional Media Forensics Methods}
Early detection efforts primarily relied on traditional media forensics techniques. These methods identify forgeries by analyzing implicit traces within digital content, with cues spanning signal-level (e.g., JPEG double compression, pixel noise patterns), physical-level (e.g., lighting inconsistencies, camera characteristics) and semantic-level (e.g., metadata anomalies). However, with the rapid development of complex generation techniques, such as GANs and diffusion models, the realism of forged content has improved. In the face of modern forgery techniques, the effectiveness of traditional methods that rely on inherent signals or physical traces~\cite{label_8,label_9,label_10,label_11} has dropped dramatically, and their limitations in complex scenes have become increasingly apparent.

\subsubsection{Deep Learning-Based Binary Classification Methods}
The rise of deep learning propelled binary classifiers to become the mainstream paradigm for deepfake detection. In image-level detection, CNNs are widely employed to analyze forgery features within single frames. For instance, Xception~\cite{label_12} enhanced feature extraction efficiency through its depthwise separable convolutions, while models based on deep architectures such as ResNeXt~\cite{label_38} focus on capturing finer-grained forgery traces. Video-level detection further incorporates temporal information, employing models that combine spatio-temporal features (such as architectures ConvNet~\cite{label_39}) to enhance detection performance. Despite the significant progress achieved by these specialized models, they often exhibit poor generalization capabilities on samples outside their training distribution, particularly those generated by unknown or novel forgery techniques. This remains a critical challenge that current research urgently needs to address.

\subsubsection{Application of Foundation Models}
Recently, foundation models have offered a new research direction for overcoming generalization challenges due to their powerful representation capabilities acquired from pretraining on large-scale data and their cross-task adaptability. The vision-language model CLIP~\cite{label_17}, through contrastive learning, initially demonstrated the ability to detect general synthetic images in zero-shot or few-shot settings~\cite{label_18}. Furthermore, self-supervised pre-trained models such as DINO~\cite{label_26} have also demonstrated excellent performance in related visual anomaly detection tasks. However, the potential of directly applying these general-purpose foundation models to the specific task of facial deepfake detection remains largely underexplored. How to effectively adapt the general capabilities of foundation models and guide their focus toward these targeted forgery features has become an important research avenue and serves as the core motivation for this study.

\section{Methodology}
While temporal consistency is a key indicator for video forgeries, our work first focuses on enhancing the generalization of single-frame analysis, as robust per-frame feature extraction is a fundamental prerequisite for effective temporal modeling. 

\begin{figure}[h] 

    \centering 

    \includegraphics[width=1\linewidth]{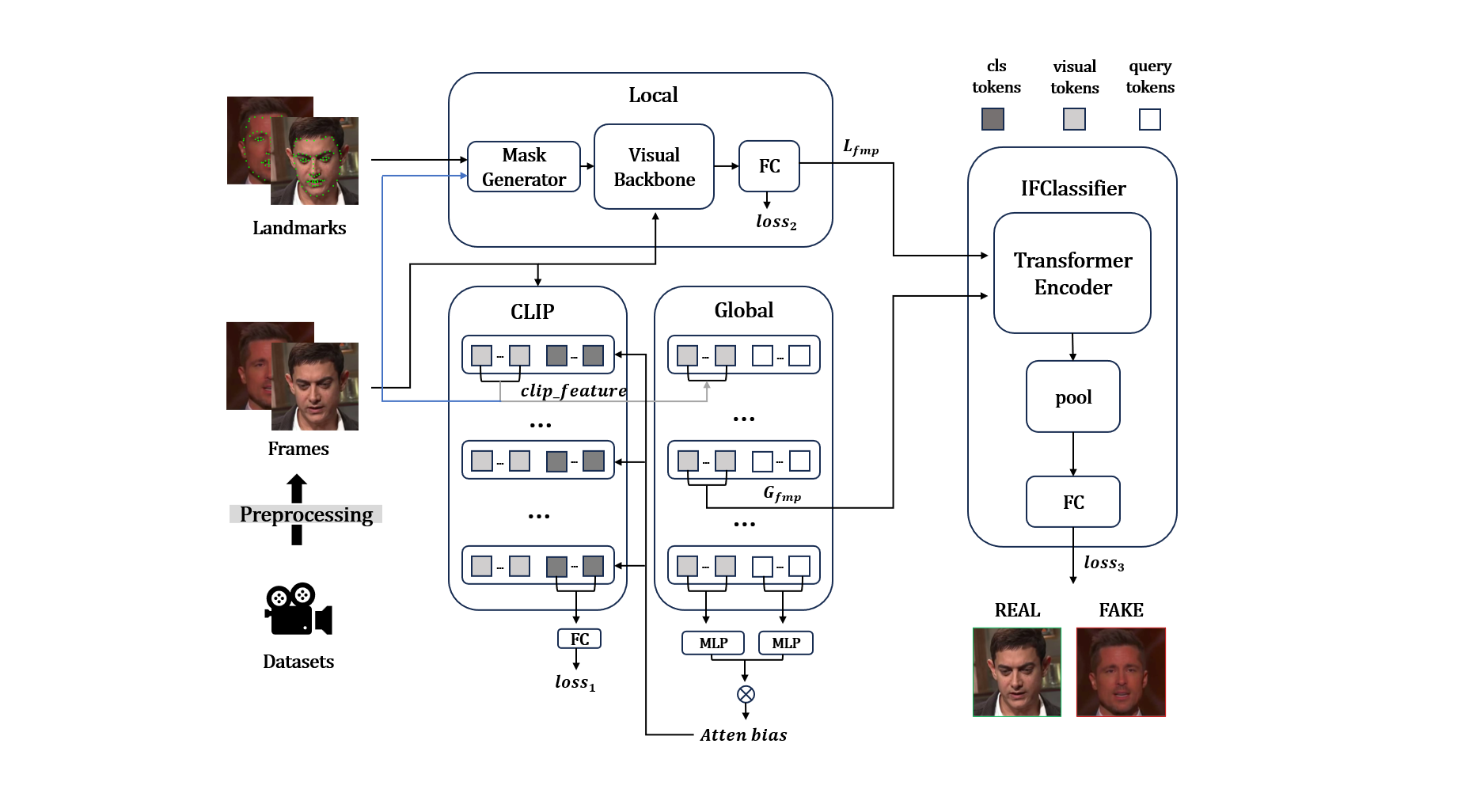} 

    \caption{\textbf{Overall architecture of the proposed Deepfake Forensics Adapter.}} 

    \label{fig:figure_3} 
\end{figure}
\subsection{Overall Framework}
Inspired by the works of~\cite{label_33,label_34}, the overall architecture of our proposed DFA framework is illustrated in Figure~\ref{fig:figure_3}.  In contrast to more general-purpose adaptation techniques like LoRA ~\cite{label_40}, Prompt Tuning ~\cite{label_41}, or standard Adapter Layers ~\cite{label_42}, our framework is explicitly designed with a dual-stream structure to simultaneously analyze global features and local facial anomalies, a critical requirement for robust deepfake detection.
Input videos first undergo preprocessing steps, including frame sampling, face detection and alignment, image normalization, and facial landmark extraction (detailed in \S\ref{sec:preprocessing_details}). The processed image frames are then fed into the frozen CLIP ViT-L/14 visual encoder to extract feature maps that are rich in semantic information. These features are passed to Global and Local stream. The Global adapter uses the query embedding to interact with the CLIP layer. It generates attention biases through a Multilayer Perceptron (MLP) and feeds them back to CLIP. This refines the global context clues related to forgery, using feature fusion to generate a global feature map $G_{fmp}$. The Local stream generates a facial region mask using landmarks information and uses an independent lightweight visual backbone  to extract a local feature map $L_{fmp}$ focused on the facial key region. Finally, the IFC module uses the Transformer encoder to deeply interact and fuse $G_{fmp}$ and $L_{fmp}$ to capture the complex dependencies between global and local forgery cues, and outputs the prediction results through the final classification head.

\subsection{Global Feature Adapter}
\subsubsection{Feature Fusion and Bias Generation}
Our Global Adapter is designed to generate a attention bias that guides the frozen CLIP model. This process involves two key stages: multi-level feature fusion from CLIP and internal bias computation.

\textbf{Multi-level Feature Fusion}
The Global Adapter has a ViT-Tiny architecture and receives visual tokens from multiple layers of the frozen CLIP encoder. To ensure compatibility, we first reshape the visual tokens from CLIP into feature maps. Specifically, features from layers $\{1, 8, 15\}$ of CLIP are processed through a $1 \times 1$ convolution to match channel dimensions and are then fused into layers $\{1, 2, 3\}$ of our adapter via element-wise addition. This multi-level fusion enriches the adapter with both low-level and high-level semantic information from the foundation model.

\textbf{Attention Bias Computation}
After processing the fused features through its initial transformer blocks, the adapter produces a set of internal query tokens $Q_{attn}$ and visual tokens $V_{attn}$. The final attention bias matrix, $B$, is computed by combining these tokens. Following a similar approach to \cite{label_34}, the bias is formulated as:
$$
B = \sum_{d=1}^{D_{out}} \text{MLP}_d(Q_{attn}) \odot \text{MLP}_d(V_{attn})
$$
Where $\odot$ denotes element-wise multiplication. $Q_{attn} \in \mathbb{R}^{N \times L_Q \times D_{in}}$ are the $L_Q$ query tokens from the adapter, and $V_{attn} \in \mathbb{R}^{N \times D_{in} \times h \times w}$ are the adapter's visual tokens, reshaped into a feature map. Both are projected by separate MLPs. The summation occurs over the output dimension $D_{out}$ of the MLPs. The resulting bias matrix $B \in \mathbb{R}^{N \times H_{attn} \times L_Q \times h \times w}$ serves as the informational bridge that interfaces with the frozen CLIP model.

\subsubsection*{Bias-Guided Attention Strategy in CLIP}
To inject the deepfake detection knowledge into CLIP without altering its weights, we employ an attention bias strategy inspired by SAN. This strategy modifies the attention mechanism's behavior by introducing shadow \texttt{[CLS]} tokens that are guided by the computed bias matrix $B$.

\textbf{Shadow Token Creation}
For each of the target CLIP layers we interact with, we first duplicate its original \texttt{[CLS]} token to create a set of shadow tokens, denoted as $X_{\text{[SLS]}}$. The full set of tokens for the attention mechanism in that layer, $X_{\text{[full]}}$, is then formed by concatenating the original visual tokens ($X_{vis}$), the original \texttt{[CLS]} token ($X_{\text{[CLS]}}$), and our new shadow tokens ($X_{\text{[SLS]}}$).

\textbf{Bias-Guided Attention Update}
The shadow tokens are updated within CLIP's self-attention layers. Their role is to actively query the full set of tokens for forgery-related information. The update rule for a shadow token at attention head $h$ in layer $l$ is defined as:
\[
X_{\text{[SLS]},h}^{l+1} = \text{Softmax}\left(\frac{Q_h(X_{\text{[SLS]}}^l) K_h(X_{\text{[full]}}^l)^T}{\sqrt{d_k}} + B\right) V_h(X_{\text{[full]}}^l)
\]
Here, $l$ denotes the layer index. The terms $Q_h(\cdot)$, $K_h(\cdot)$, and $V_h(\cdot)$ represent the query, key, and value vectors at a specific attention head $h$. The query is generated from the shadow tokens $X_{\text{[SLS]}}^l$, while the key and value are generated from the full set of tokens $X_{\text{[full]}}^l$. The dimension of the key vector is $d_k$.

\subsection{Local Anomaly Stream}
The Local stream is designed to enhance perception of local forgery cues in critical regions (e.g.,  eyes, mouth, nose).  It leverages facial structure priors to detect anomalies and inconsistencies often introduced by facial forgery, a strategy supported by research demonstrating the effectiveness of using spatial and temporal features derived from facial landmarks for robust detection~\cite{label_43}.

The Local stream receives the image frames, $clip\_feature$, and corresponding facial landmark coordinates as input. A Landmark Mask Generator uses the input 81 landmarks and $clip\_feature$, according to predefined facial contents (e.g., eyebrows, eyes, nose, lips), to generate corresponding spatial attention masks for each region. The Local stream also employs a Convolutional Neural Network (CNN) independent of CLIP as its visual feature extraction backbone. This backbone processes the input image to learn and extract feature maps $L_{fmp}$ highly relevant to facial regions. Similar to the Global Feature Adapter, the Local stream incorporates an auxiliary classification head. This head performs an independent authenticity prediction based on the features extracted by the Local stream, providing an additional supervisory signal $loss_2$ for model training.

\subsection{Interactive Fusion Classifier}
The IFC integrates the global features from the Global adapter with local details from the Local stream. It uses a transformer encoder to fuse these complementary feature streams, capturing complex dependencies between them to form a comprehensive forgery representation for the final classification.

The IFC module concatenates the input $G_{fmp}$ and $L_{fmp}$ along the sequence dimension, uses its internal transformer encoder for deep interaction and fusion, and generates the final prediction after processing through the pool layer and the fully connected layer.

\subsection{Training Objective}
To ensure effective learning of the DFA framework, we adopt a multitask learning paradigm, simultaneously supervising the predictions from the Global adapter, the Local stream and the IFC module. 
The overall optimization objective is to minimize a combined loss function:
\begin{equation} \label{eq:total_loss} 
L_{total} = w_{global} \cdot loss_1 + w_{local} \cdot loss_2 + w_{fusion} \cdot loss_3
\end{equation}
Where $w_{global}$, $w_{local}$, and $w_{fusion}$ are the respective weight coefficients. 
Notably, we define these weights as learnable parameters.

\section{Experiments}
This chapter aims to comprehensively evaluate the effectiveness of the proposed DFA method through a series of experiments. We will detail the datasets employed, the data preprocessing pipeline, key details of the model implementation, and the performance evaluation metrics used. Furthermore, we will present and analyze the detailed experimental results.
\subsection{Datasets}
\begin{figure}[htbp] 

    \centering 

    \includegraphics[width=1\linewidth]{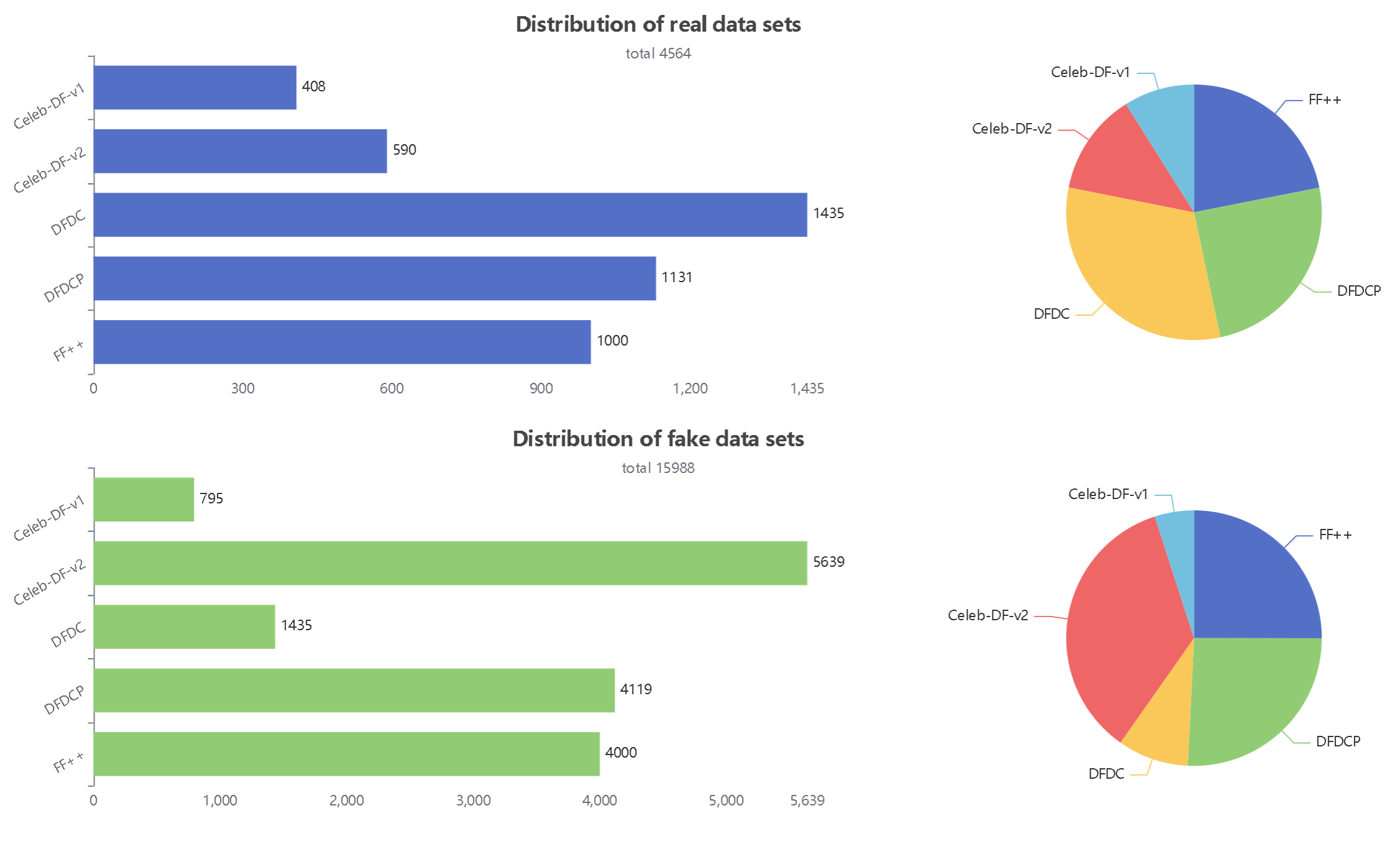} 

    \caption{ Overview of datasets used for evaluating model performance, including CelebDF-V1, CelebDF-V2, DFDC, DFDCP, and FF++. DFDC is reserved for evaluating generalization capability, while the remaining datasets form the training/validation data.} 

    \label{fig:figure_4} 
\end{figure}

For a comprehensive model evaluation, we use five widely-used public facial forgery datasets: Celeb-DF-v1, Celeb-DF-v2~\cite{label_27}, Deepfake Detection Challenge (DFDC), DFDCP~\cite{label_28}, and FaceForensics++ (FF++)~\cite{label_12}. The specific composition and distribution of these datasets are summarized in Figure~\ref{fig:figure_4}.

To rigorously evaluate the model's generalization capability, we reserve the DFDC dataset, which is considered to be extremely challenging, as an independent test set that remains unseen by the model during the training phase. The remaining four datasets (CelebDF-v1, CelebDF-v2, DFDCP, and FF++) are combined to form a Mixed Dataset. This Mixed Dataset is then strictly partitioned into training and validation/test sets using an 80\%:20\% ratio. These subsets are used for model parameter learning and performance monitoring during training and preliminary evaluation, respectively.
\subsection{Data Preprocessing}
\label{sec:preprocessing_details}
To ensure the reproducibility of our experimental results and fairness in cross-dataset comparisons, we strictly adhered to the standardized data processing pipeline proposed by DeepfakeBench~\cite{label_29}, applying uniform preprocessing steps across all datasets used. This process commences with sampling video frames from each original video, which can be achieved either by extracting a fixed number of frames (e.g., 32 frames were selected by default in this study) or by sampling at a fixed time stride. Subsequently, for each extracted frame, face detection and alignment are performed using the Dlib library. Based on the shape predictor model provided by Dlib, we precisely extracted 81 facial landmarks for each detected face. Finally, image normalization is conducted: images are cropped according to the detected face bounding boxes, and all resulting facial images are uniformly resized to a standard resolution of 256×256 pixels. Concurrently, the coordinates of the extracted facial landmarks undergo a corresponding normalization transformation to match the adjusted image dimensions. This series of standardized preprocessing steps provides regularized and consistent data input for subsequent model training and evaluation.
\subsection{Implementation Details}
Our experiments are based on the pre-trained CLIP ViT-L/14 visual encoder~\cite{label_17} with its parameters frozen. 
The input image resolution is set uniformly to $224 \times 224$ pixels. 
Facial landmarks are provided in the form of a tensor with shape $[batchsize, 81, 2]$.
The Visual Backbone within the Local stream uses a ResNeXt-50 architecture, with its final two layers removed. Model training uses the Adam optimizer with an initial learning rate set to $2 \times 10^{-6}$. 
All experiments were carried out on a system equipped with 4 NVIDIA Tesla V100S GPUs, and the random seed was set to 706. Training was performed for a total of 6 epochs with a batch size of 32, using 64 CPU workers for data loading. The coefficients $w_{global}$, $w_{local}$, and $w_{fusion}$ used to weight the different loss terms within the framework are configured as learnable parameters. 
The overall training duration is approximately 10 GPU hours.
\subsection{Evaluation Metrics}
\label{sec:eval_metrics}
To comprehensively evaluate, from multiple perspectives, the performance of our model on the binary classification task of distinguishing between Real and Fake samples, we employ the following four widely used evaluation metrics: Accuracy, Precision, Area Under the ROC Curve (AUC), and Equal Error Rate (EER).
Before detailing these metrics, we first define the basic terms derived from the confusion matrix:
\begin{itemize}
    \item \textbf{True Positive (TP)}: A Fake sample correctly predicted as Fake. 
    \item \textbf{True Negative (TN)}: A Real sample correctly predicted as Real. 
    \item \textbf{False Positive (FP)}: A Real sample incorrectly predicted as Fake. 
    \item \textbf{False Negative (FN)}: A Fake sample incorrectly predicted as Real. 
\end{itemize}
Based on these definitions, the evaluation metrics are described as follows:
\begin{description}
    \item[\textbf{Accuracy}] Measures the overall proportion of correct predictions (both TP and TN) across all samples. It reflects the model's general classification accuracy on the entire dataset. The formula is $Accuracy = \frac{TP + TN}{TP + TN + FP + FN}$.

    \item[\textbf{Precision}] Refers to the proportion of samples predicted as Fake (positive) by the model that are actually Fake. This metric focuses on the reliability of the positive predictions; high precision indicates that the model's predictions of "Fake" are more trustworthy. The formula is $Precision = \frac{TP}{TP + FP}$.

    \item[\textbf{Area Under the ROC Curve (AUC)}] The AUC value quantifies the overall ability of the models to distinguish between positive and negative samples across all possible classification thresholds. The Receiver Operating Characteristic (ROC) curve is plotted with the False Positive Rate (FPR) on the x-axis and the True Positive Rate on the y-axis. AUC values range between 0 and 1, where a value closer to 1 indicates better discrimination performance, and 0.5 typically represents random guessing. The formulas for TPR and FPR are $TPR = \frac{TP}{TP + FN}$ and $FPR = \frac{FP}{FP + TN}$. 

    \item[\textbf{Equal Error Rate (EER)}] EER corresponds to the error rate at the specific operating point on the ROC curve where the False Positive Rate (FPR) equals the False Negative Rate (FNR), i.e., $FPR = FNR$. 
    A lower EER value indicates better optimal performance that the model can achieve in balancing false alarms (FP) and missed detections (FN), signifying stronger overall discrimination ability. The formula for FNR is $FNR = \frac{FN}{FN + TP}$.
\end{description}
\subsection{Results and analysis}
This section reports and analyzes the performance of our proposed DFA on multiple benchmark datasets, comparing it with existing representative methods.
All experiments are evaluated on the mixed dataset and the DFDC dataset at both the frame-level and video-level, using the Accuracy, Precision, AUC, and EER metrics detailed in Section~\ref{sec:eval_metrics}. 
To ensure experimental rigor and comparability of results, unless otherwise specified, all comparison experiments adhere to the following conventions:
\begin{enumerate}
    \item All models included in the comparison are trained using the training subset of the aforementioned Mixed dataset;
    \item In the reported experimental results, all metric values are presented with three decimal places; 
    \item In all results tables, bold values indicate the best performance for the corresponding metric, while underlined values indicate the second-best performance.
\end{enumerate}

\begin{table}[htbp] 
    \centering 
    \caption{\textbf{Frame-level} metrics scores of various deepfake detection methods on our Mixed dataset.} 
    \renewcommand{\arraystretch}{1.5}
    \label{tab:table_1} 
    \begin{tabular}{@{} p{3.5cm}>{\centering\arraybackslash}p{2cm} >{\centering\arraybackslash}p{2cm} >{\centering\arraybackslash}p{2cm} >{\centering\arraybackslash}p{1cm}@{}}
        \toprule
        Method           & Acc   & Precision & AUC   & Avg   \\
        \midrule 
        CNN~\cite{label_30}        & 0.904 & 0.815 & 0.890 & 0.867 \\
        Xception~\cite{label_12}    & 0.955 & 0.897 & \underline{0.962} & 0.938 \\
        ResNeXt-LSTM~\cite{label_14}  & 0.967 & 0.931 & 0.960 & \underline{0.953} \\
        CViT~\cite{label_13}    & 0.941 & 0.868 & 0.959 & 0.923 \\
        Efficient-ViT~\cite{label_15}& \underline{0.968} & \underline{0.932} & 0.954 & 0.951 \\
        Ours            & \textbf{0.983} & \textbf{0.963} & \textbf{0.976} & \textbf{0.974} \\ 
        \bottomrule 
    \end{tabular}
\end{table}

\subsubsection{Performance on the Mixed Dataset}
As shown in Table~\ref{tab:table_1}, our method achieved the best frame-level performance on the mixed dataset across all metrics. Notably, its Accuracy, Precision, and AUC scores were 1.5\%, 3.1\%, and 1.4\% higher than the next-best method, respectively.

\subsubsection{Frame-level Performance on DFDC Dataset} 
As shown in Table~\ref{tab:table_2}, DFA achieved a SOTA frame-level AUC of 0.816, significantly surpassing the second-best method (Efficient-ViT at 0.764). It also recorded the lowest EER of 0.256 among all evaluated methods.

\begin{table}[h] 
    \centering 
    \caption{\textbf{Frame-level} AUC and EER scores of various deepfake detection methods on the DFDC dataset.} 
    \label{tab:table_2}
    \renewcommand{\arraystretch}{1.5}
    \begin{tabular}{@{} p{3.5cm} >{\centering\arraybackslash}p{2cm} >{\centering\arraybackslash}p{2cm} @{}}
        \toprule 
        Method           & AUC   & EER$\downarrow$ \\ 
        \midrule 
        CNN~\cite{label_30}         & 0.674 & 0.370 \\
        Xception~\cite{label_12}    & 0.752 & 0.315 \\
        ResNeXt-LSTM~\cite{label_14}  & 0.754 & 0.311 \\
        CViT~\cite{label_13}    & 0.738 & 0.323 \\
        Efficient-ViT~\cite{label_15}& \underline{0.764} & \underline{0.293} \\
        Ours            & \textbf{0.816} & \textbf{0.256} \\ 
        \bottomrule 
    \end{tabular}
\end{table}

\subsubsection{Video-level Performance on DFDC Dataset} 
Considering practical application scenarios for video detection, we further evaluated video-level performance on the DFDC dataset, with results shown in Table 3. DFA again demonstrates a leading advantage, achieving the best video-level AUC of 0.836. This represents a substantial 4.8\% improvement over the second-best method, Efficient-ViT (0.788). Additionally, DFA obtained the lowest video-level EER of 0.251. These consistent results across both frame and video levels strongly validate the excellent generalization performance of the DFA model when confronted with unseen data.

\begin{table}[h] 
    \centering 
    \caption{\textbf{Video-level} AUC and EER scores of various deepfake detection methods on the DFDC dataset.} 
    \label{tab:table_3} 
    \renewcommand{\arraystretch}{1.5}
    \begin{tabular}{@{} p{3.5cm} >{\centering\arraybackslash}p{2cm} >{\centering\arraybackslash}p{2cm} @{}}
        \toprule 
        Method           & AUC   & EER$\downarrow$ \\ 
        \midrule 
        CNN~\cite{label_30}        & 0.698 & 0.375 \\
        Xception~\cite{label_12}    & 0.774 & 0.325 \\
        ResNeXt-LSTM~\cite{label_14}  & 0.776 & 0.319 \\
        CViT~\cite{label_13}    & 0.770 & 0.323 \\
        Efficient-ViT~\cite{label_15}& \underline{0.788} & \underline{0.300} \\
        Ours            & \textbf{0.836} & \textbf{0.251} \\
        \bottomrule
    \end{tabular}
\end{table}

\begin{table}[htbp] 
    \centering 
    \caption{Ablation study on DFA modules: \textbf{Frame-level} results on the DFDC dataset.} 
    \label{tab:table_4} 
    \renewcommand{\arraystretch}{1.5}
    \begin{tabular}{@{} p{2cm}>{\centering\arraybackslash}p{2cm} >{\centering\arraybackslash}p{2cm} >{\centering\arraybackslash}p{2cm} >{\centering\arraybackslash}p{1cm}@{}}
        \hline 
        Global & Local & IFC & AUC & EER$\downarrow$ \\ 
        \hline 
        $\times$     & $\checkmark$ & $\checkmark$ & \underline{0.766}          & 0.356          \\ 
        $\checkmark$ & $\times$     & $\checkmark$ & 0.747          & 0.323          \\
        $\checkmark$ & $\checkmark$ & $\times$     & 0.753          & \underline{0.301}          \\
        $\checkmark$ & $\checkmark$ & $\checkmark$ & \textbf{0.816} & \textbf{0.256} \\ 
        \hline 
    \end{tabular}
\end{table}

\subsubsection{Ablation Study}
To validate the effectiveness and necessity of each key component within our proposed DFA framework, we performed a series of ablation studies on the DFDC dataset. The experiments assess the impact on overall model performance, measured by frame-level AUC and EER, by progressively removing or disabling specific modules. The experimental results are summarized in Table~\ref{tab:table_4}, and its ablation study verifies the contribution of each module. When the Global module is ablated, AUC drops significantly from 0.816 to 0.766, and EER rises from 0.256 to 0.356, highlighting its critical function in using CLIP features for discrimination. Performance also markedly decreases upon removing the other components: ablating the Local stream yields an AUC of 0.747 and an EER of 0.323, while ablating the IFC module results in an AUC of 0.753 and an EER of 0.301. These results confirm that the Global, Local, and IFC modules each play an indispensable role.

\subsubsection{T-SNE Visualization}
To more intuitively evaluate the discriminative power of the feature representations learned by our proposed DFA model, we used the t-SNE~\cite{label_31} algorithm for dimensionality reduction and visualization of the features extracted by the model from samples in the DFDC test set. Simultaneously, we compared these visualizations with those of features extracted by the baseline method, Xception~\cite{label_12}. For this experiment, we randomly sampled 500 real images and 500 fake images from the DFDC dataset. For our DFA model, the visualization employs the final fused features processed by the IFC module. For the Xception model, the features obtained after its global average pooling layer are used.

\begin{figure}[htbp] 

    \centering 

    \includegraphics[width=0.8\linewidth]{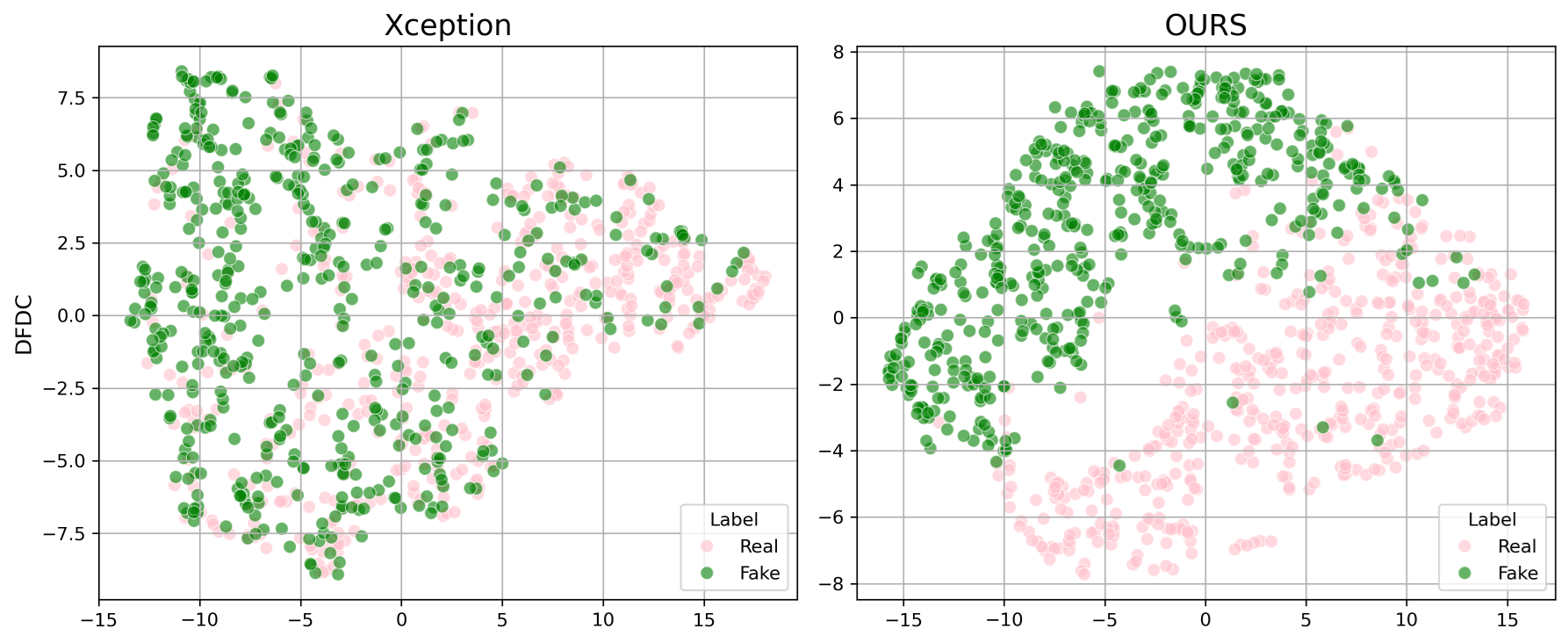} 

    \caption{Comparison of t-SNE visualizations of features extracted by Xception (left) and our DFA model (right) on samples from the DFDC dataset.} 

    \label{fig:figure_5} 
\end{figure}

The visualization results, shown in Figure~\ref{fig:figure_5}, clearly demonstrate that compared to Xception, our DFA model more effectively maps the features of real and fake samples into distinct and more clearly separated clusters within the feature space. This intuitively confirms that the DFA model has stronger feature learning and discriminative capabilities.

\section{Conclusion}
To address the limited generalization capability of current deepfake detection methods when faced with various unknown forgery techniques, we propose a novel framework named Deepfake Forensics Adapter (DFA). This framework innovates by achieving effective interaction with and adaptation to the powerful CLIP vision encoder through a customized adapter module, while keeping CLIP's parameters frozen. Key components include a Local stream that leverages facial structural information to extract local features, and an IFC module responsible for multisource feature fusion. We conducted comprehensive experiments on a mixed dataset that includes diverse forgery techniques, as well as on the highly challenging DFDC benchmark dataset. The results demonstrate DFA's superior performance across multiple metrics. Notably, on the training-unseen DFDC dataset, DFA achieved state-of-the-art performance in both frame-level and video-level detection. It achieved the highest frame-level AUC score of 0.816 and a video-level AUC score of 0.836, along with the lowest frame-level EER of 0.256 and video-level EER of 0.251. The video-level AUC notably surpassed the second-best baseline method by a significant 4.8\% margin. These findings strongly demonstrate that effectively adapting large pre-trained models via carefully designed adapters represents a viable and effective approach for enhancing the generalization of 
deepfake detection. This research contributes a thoroughly experimentally validated state-of-the-art solution to address the increasingly sophisticated deepfake threats.

\section{Limitations and Future Work}
While our proposed DFA framework demonstrates a leading performance in the deepfake detection task, there remains room for further improvement. Current limitations are primarily twofold: (1) Although we adopt the frame sampling strategy used in the current benchmark to standardize evaluation,  this approach restricts the full utilization of long-range temporal dynamic information in videos.(2) As this study focuses exclusively on facial forgeries, the effectiveness of our approach against content generated by emerging techniques (e.g., diffusion models) or other types of AI-generated content (e.g., full-body forgeries, audio-video manipulations) warrants further investigation.

Moving forward, we aim to integrate advanced temporal modeling mechanisms (e.g., temporal convolutions, recurrent networks, or temporal attention mechanisms) into the adapter framework to better exploit dynamics in videos, while also extending DFA's adaptation methodology to broader detection contexts including evaluation against state-of-the-art generative models and multimodal forgery scenarios.

\end{document}